\documentclass{article}
\pdfoutput=1
\usepackage{times}
\usepackage{subfigure} 
\usepackage{natbib}

\usepackage{graphicx}
\usepackage{mathtools} 
\usepackage{amsthm, amssymb, amscd, amsfonts} 
\usepackage[ampersand]{easylist} 
\usepackage{multirow} 
\usepackage{booktabs} 
\usepackage{bm} 
\usepackage{dcolumn} 
\usepackage{wrapfig}
\usepackage{bbm}
\usepackage{verbatim}
\usepackage{sqr-notation}

\usepackage[accepted]{icml2016}
\icmltitlerunning{Square Root Graphical Models}
\begin{document}
\twocolumn[
\icmltitle{
Square Root Graphical Models: Multivariate Generalizations of \\
Univariate Exponential Families that Permit Positive Dependencies}
\icmlauthor{David I. Inouye}{dinouye@cs.utexas.edu}
\icmlauthor{Pradeep Ravikumar}{pradeepr@cs.utexas.edu}
\icmlauthor{Inderjit S. Dhillon}{inderjit@cs.utexas.edu}
\icmladdress{Dept. of Computer Science, University of Texas, Austin, TX 78712, USA}
\icmlkeywords{Graphical Models, Exponential Family Distributions, Poisson Graphical Models, Exponential Graphical Models, Gaussian Graphical Models}
\vskip 0.3in
]


\begin{abstract}
We develop Square Root Graphical Models (SQR), a novel class of parametric graphical models that provides multivariate generalizations of univariate exponential family distributions.  Previous multivariate graphical models \citep{Yang2015} did not allow positive dependencies for the exponential and Poisson generalizations. However, in many real-world datasets, variables clearly have positive dependencies.  For example, the airport delay time in New York---modeled as an exponential distribution---is positively related to the delay time in Boston. With this motivation, we give an example of our model class derived from the univariate exponential distribution that allows for almost arbitrary positive and negative dependencies with only a mild condition on the parameter matrix---a condition akin to the positive definiteness of the Gaussian covariance matrix. Our Poisson generalization allows for both positive and negative dependencies without any constraints on the parameter values. We also develop parameter estimation methods using node-wise regressions with $\ell_1$ regularization and likelihood approximation methods using sampling. Finally, we demonstrate our exponential generalization on a synthetic dataset and a real-world dataset of airport delay times.
\end{abstract}

\section{Introduction}
Gaussian, binary and discrete undirected graphical models---or Markov Random Fields (MRF)---have become popular for compactly modeling and studying the structural dependencies between high-dimensional continuous, binary and categorical data respectively \citep{Friedman2008,Hsieh2014,Banerjee2008a,Ravikumar2010,Jalali2010}.  However, real-world data does not often fit the assumption that variables come from Gaussian or discrete distributions. For example, word counts in documents are nonnegative integers with many zero values and hence are more appropriately modeled by the Poisson distribution. Yet, an independent Poisson distribution  would be insufficient because words are often either positively or negatively related to other words---e.g. the words ``machine'' and ``learning'' would often co-occur together in ICML papers (positive dependency) whereas the words ``deep'' and ``kernel'' would rarely co-occur since they usually refer to different topics (negative dependency). Thus, a Poisson-like model that allows for dependencies between words is desirable. As another example, the delay times at airports are nonnegative continuous values that are more closely modeled by an exponential distribution than a Gaussian distribution but an independent exponential distribution is insufficient because delays at different airports are often related (and sometimes causally related)---e.g. if a flight from Los Angeles, CA (LAX) to San Francisco, CA (SFO) is delayed then it is likely that the return flight of the same airplane will also be delayed. Other examples of non-Gaussian and non-discrete data include high-throughput gene sequencing count data, crime statistics, website visits, survival times, call times and delay times.

Though univariate distributions for these types of data have been studied quite extensively, multivariate generalizations have only been given limited attention.  One basic approach to forming dependent multivariate distributions is to assume that the marginal distributions are exponentially distributed \citep{Marshall1967,Embrechts2003} or Poisson distributed \citep{Karlis2003}.  This idea is related to copula-based models \citep{Bickel2009a} in which a probability distribution is decomposed into the univariate marginal distributions and a copula distribution on the unit hypercube that models the dependency between variables. However, the exponential model in \citep{Marshall1967,Embrechts2003} gives rise to a distribution that is composed of a continuous distribution and a singular distribution, which seems unusual and unlikely for general real-world situations.  The multivariate Poisson distribution \citep{Karlis2003} is based on the sum of independent Poisson variables and can only model \emph{positive} dependencies. The copula versions of the multivariate Poisson distribution have significant issues related to non-identifiability because the Poisson distribution has a discrete domain \citep{Genest2007}.  There has also been some recent work on semi-parametric graphical models \citep{Liu2009a} that use Gaussian copulas to relax the assumption of Gaussianity but these models are not parametric and only consider continuous real-valued data.

Another line of work assumes that the node conditional distributions---i.e. one variable given the values of all the other variables---are univariate exponential families\footnote{See \citep{Wainwright2007} for an introduction to exponential families.} and determines under what conditions a joint distribution exists that is consistent with these node conditional distributions.    \citet{Besag1974} developed this multivariate distribution for pairwise dependencies, and \citet{Yang2015} extended this model to n-wise dependencies.  \citet{Yang2015} also developed and analyzed an M-estimator based on $\ell_1$ regularized node-wise regressions to recover the graphical model structure with high probability. Unfortunately, these models only allowed \emph{negative} dependencies in the case of the exponential and Poisson distributions.  \citet{Yang2013} proposed three modifications to the original Poisson model to allow positive dependencies but these modifications alter the Poisson base distribution or require the specification of unintuitive hyperparameters.  \citet{Allen2013} allowed positive dependencies by only requiring the Local Markov property rather than a consistent joint distribution that would have Global Markov properties.  In a different approach, \citet{Inouye2015} altered the Poisson generalization by assuming the length of the vector is fixed or known similar to the multinomial distribution in which the number of trials is known.  This allows a joint distribution that is decomposed into the marginal distribution of vector length and the distribution of the vector direction given the length.  While the model in \citep{Inouye2015} allowed for both positive and negative dependencies, the joint distribution needed to be modified by an ad hoc scalar weighting function to avoid very low likelihood values for vectors of long length---i.e. documents with many words.

Therefore, we develop a novel parametric generalization of univariate exponential family distributions with nonnegative sufficient statistics---e.g. Gaussian, Poisson and exponential---that allows for \emph{both positive and negative dependencies}.  We call this novel class of multivariate distributions Square Root Graphical Models (SQR) because the square root function is fundamentally important as will be described in future sections.  SQR models have a simple parametric form without needing to specify any hyperparameters and can be fit using $\ell_1$-regularized node-wise regressions similar to previous work \citep{Yang2015}.  The independent model---e.g. independent Poisson or exponential---is merely a special case of this class unlike in \citep{Yang2013}. We show that the normalizability of the distribution puts little to no restriction on the values of the parameters, and thus SQR models give a very flexible multivariate generalization of well-known univariate distributions.

\paragraph{Notation} Let $\vmax$ and $\imax$ denote the number of dimensions and number data instances respectively.  We will generally use uppercase letters for matrices (e.g. $\pmat, \instmat$), boldface lowercase letters for vectors (i.e $\instvec, \pvec$) and lowercase letters for scalar values (i.e. $\inst, \psca$). Let $\Rp$ denote the set of nonnegative real numbers and $\Zp$ denote the set of nonnegative integers.

\section{Background}
To motivate the form of our model class, we present a brief background on the graphical model class as in \citep{Besag1974,Yang2015,Yang2013}. Let $\T(\inst)$  and $\B(\inst)$ be the sufficient statistics and log base measure respectively of the base univariate exponential family and let $\domain \subseteq \Rp^\vmax$ be the domain of the random vector.  We will denote $\T(\instvec)\colon \R^\vmax \to \R^\vmax$ to be the entry-wise application of the sufficient statistic function to each entry in the vector $\instvec$.  With this notation, the previous class of graphical models can be defined as \cite{Yang2015}:
\begin{align}
&\begin{array}{l}
\Pr(\instvec | \theta, \pmat) = \exp\Big( \bm{\theta}^T \T(\instvec) + { \T(\instvec)^T \pmat \T(\instvec)} \\
\hspace{9em}  + \sum_{\vi=1}^{\vmax} \B(\inst_\vi) - \A(\bm{\theta},\pmat) \Big)
\end{array} \label{eqn:glm-gm} \\
&\begin{array}{l}
\A(\bm{\theta},\pmat) = \int_{\domain} \exp\Big( \bm{\theta}^T \T(\instvec) + {\T(\instvec)^T \pmat \T(\instvec)} \\
\hspace{9em} + \sum_{\vi=1}^{\vmax} \B(\inst_\vi) \Big) \mathrm{d}\measure(\instvec) \, ,
\end{array}\label{eqn:glm-gmA}
\end{align}
where $\A(\bm{\theta},\pmat)$ is the log partition function (i.e. log normalization constant) which is required for probability normalization, $\pmat \in \R^{\vmax \times \vmax}$ is symmetric with zeros along the diagonal and $\measure$ is either the standard Lebesgue measure or the counting measure depending on whether the domain $\domain$ is continuous or discrete.  The only difference from a fully independent model is the quadratic interaction term $\T(\instvec)^T \pmat \T(\instvec)$---i.e. $O(\T(\inst)^2)$---which is why the exponential and Poisson cases do not admit positive dependencies as will be described in later sections.

We will review the exponential instantiation of this previous graphical model in which  the domain $\domain \in \Rp^\vmax$, $\T(\inst) = \inst$ and $\B(\inst) = 0$.  Suppose there is even one positive entry in $\pmat$ denoted $\psca_{\vi\vit}$. Then as $\instvec \to \infty$, the positive quadratic term $\inst_\vi\psca_{\vi\vit}\inst_\vit$ will dominate the linear term $\bm{\theta}^T \instvec$ and thus the log partition function will diverge (i.e. $\A(\bm{\theta},\pmat) \to \infty$).  Thus, $\pmat_{\vi\vit} \leq 0$ is required for a consistent joint distribution. Similarly, in the case of the Poisson distribution where the domain $\domain \in \Zp^\vmax$, $\T(\inst) = \inst$ and $\B(\inst) = -\log(\inst!)$, suppose there is even one positive entry $\psca_{\vi\vit}$.  The quadratic term $\inst_\vi\psca_{\vi\vit}\inst_\vit$ will dominate the linear term \emph{and} the log base measure term which is $O(\inst \log(\inst))$; thus, $\pmat_{\vi\vit} \leq 0$ is also required for the Poisson distribution.

In an attempt to allow positive dependencies for the Poisson distribution, \citet{Yang2013} developed three variants of the Poisson graphical model defined above.  First, they developed a Truncated Poisson Graphical Model (TPGM) that kept the same parametric form but merely truncated the usual infinite domain to the finite domain $\domain_{\text{TPGM}} = \{ \instvec \in \Zp^\vmax: \inst_\vi \leq \tpgmR \}$. However, a user must a priori specify the truncation value $\tpgmR$ and thus TPGM is unnatural for normal count data that could be infinite.  In addition, because of the quadratic term, even though the domain is finite, the quadratic term can dominate and push most of the mass near the boundary of the domain \citep{Yang2013}.  The second proposal was to change the base measure from $\log(\inst!)$ to $\inst^2$.  This proposal, however, gives the distribution Gaussian-like quadratic tails rather than the thicker tails of the Poisson distribution.  Finally, the last proposal modified the sufficient statistic $\T(\inst)$ to decrease from linear to constant as $\inst$ increases.  Similar to TPGM, this third proposal requires the a priori specification of two cutoff parameters $(\spgmR_1, \spgmR_2)$ and behaves similarly to TPGM after the second cutoff point because the base measure of $-\log(\inst!)$ will quickly make the probability approach 0 once the sufficient statistics become constant.

In a somewhat different direction, \citet{Inouye2015} proposed a variant called Fixed-Length Poisson MRF (LPMRF) that modifies the domain of the distribution assuming the length of the vector $\length = \|\instvec\|_1$ is fixed, i.e. $\domain = \{ \instvec \in \Zp^\vmax : \|\instvec\|_1 = \length \}$.  Because the domain is finite as in TPGM, the distribution is normalizable even with positive dependencies.  However, as with TPGM, the quadratic term in the parametric form dominates the distribution if $\length$ is large, and thus \citet{Inouye2015} modify the distribution by introducing a weighting function that decreases the quadratic term as $\length$ increases.  All of these variants of the Poisson graphical model attempt to deal with the quadratic interaction term in different ways but all of them significantly change the distribution/domain and often require the specification of new unintuitive hyperparameters to allow for positive dependencies. Also, according to the authors' best knowledge, no variants of the exponential graphical model have been proposed to allow for positive dependencies. Therefore, we propose a novel graphical model class that alleviates the problem with the quadratic interaction term and provides both exponential and Poisson graphical models that allow \emph{positive and negative} dependencies.

\section{Square Root Graphical Model}
\begin{figure*}[!ht]
\centering
\includegraphics[width=0.8\textwidth]{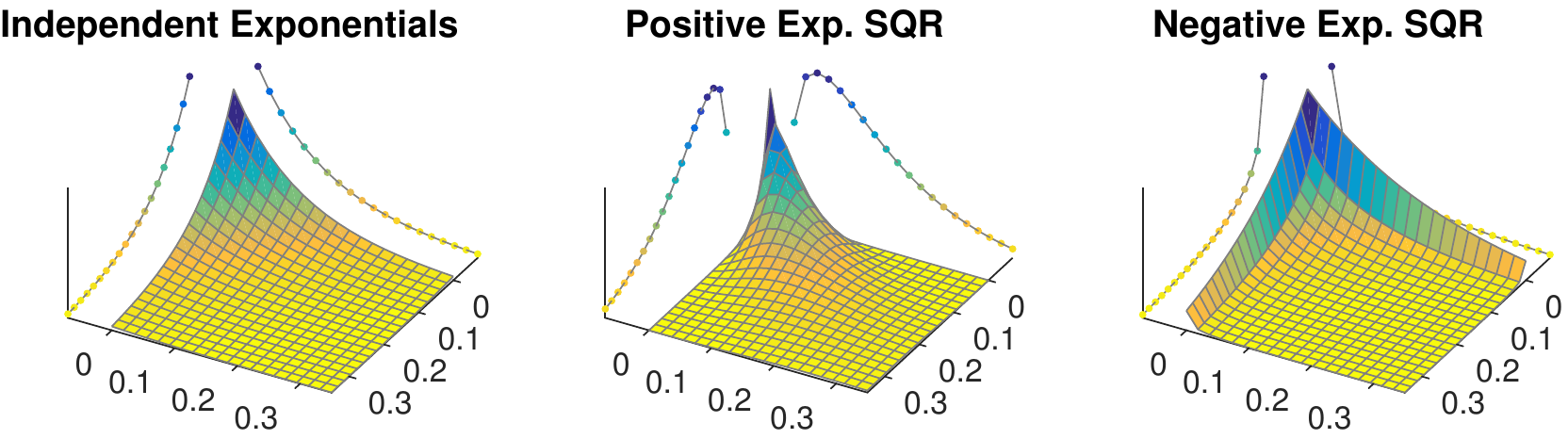} \\
\vspace{0.25em}
\includegraphics[width=0.8\textwidth, trim=1.95cm 0 1.7cm 0, clip]{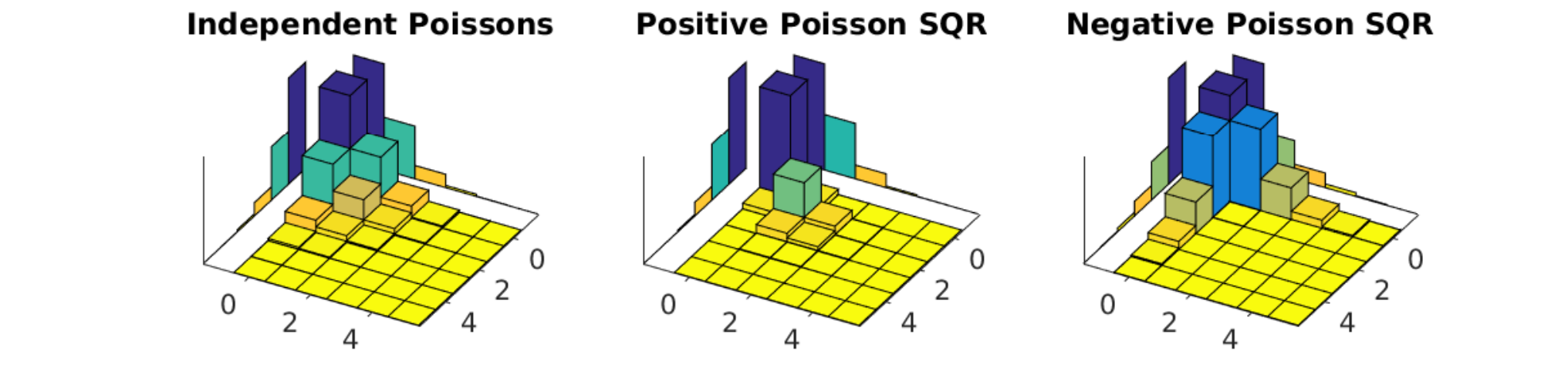}
\caption{These examples of 2D exponential SQR and Poisson SQR distributions with no dependency (i.e. independent), positive dependency and negative dependency show the amazing flexibility of the SQR model class that can intuitively model \emph{positive and negative} dependencies while having a simple parametric form. The approximate 1D marginals are shown along the edges of the plots.}
\label{fig:marginals}
\end{figure*}
\vspace{-0.25em}

The amazingly simple yet helpful change from the previous graphical model class in Eqn.~\ref{eqn:glm-gm} is that we take the square root of the sufficient statistics in the interaction term.  Essentially, this makes the interaction term linear in the sufficient statistics $O(\T(\inst))$ rather than quadratic $O(\T(\inst)^2)$ as in Eqn.~\ref{eqn:glm-gm}. This change avoids the problem of the quadratic term overcoming the other terms while allowing both positive and negative dependencies. More formally, given any univariate exponential family with nonnegative sufficient statistics $\T(\inst) \geq 0$, we can define the Square Root Graphical Model (SQR) class as follows:
\begin{align}
&\begin{array}{l}
\Pr(\instvec \,|\, \nodepvec, \pmat) \!=\! \exp\!\Big(\! \nodepvec^T \!{\sqrt{\T(\instvec)} } \!+\! { \sqrt{\T(\instvec)}^T \!\pmat \sqrt{\T(\instvec)} }\\
\hspace{11em} + \sum_\vi \!\B(\inst_\vi) - \A(\pmat)\Big)
\end{array} \label{eqn:sqr-model} \\
&\begin{array}{l}
\A(\nodepvec, \pmat) \!=\! \int_{\domain} \exp\!\Big(\!\nodepvec^T\!{\sqrt{\T(\instvec)} } \!+\! { \sqrt{\T(\instvec)}^T \!\pmat \sqrt{\T(\instvec)} } \\
\hspace{10em} + \sum_\vi \! \B(\inst_\vi)\Big) \mathrm{d}\measure(\instvec) \, ,
\end{array} \label{eqn:sqr-modelA}
\end{align}
where $\sqrt{\T(\instvec)}$ is an entry-wise square root except when $\T(\inst) = \inst^2$ in which case $\sqrt{\T(\inst)} \equiv \inst$.\footnote{This nuance is important for the Gaussian SQR in Sec.~\ref{sec:gaussian-sqr}.} Figure~\ref{fig:marginals} shows examples of the exponential and Poisson SQR distributions for no dependency, positive dependency and negative dependency. If $\nodepvec = 0$ and $\pmat$ is a diagonal matrix, then we recover an independent joint distribution so the SQR class of models can be seen as a direct relaxation of the independence assumption, similar to previous graphical models.  In the next sections, we analyze some of the properties of SQR models including their conditional distributions.

\subsection{SQR Conditional Distributions}
\label{sec:conditionals}
\begin{figure}[!ht]
\centering
\includegraphics[width=\columnwidth]{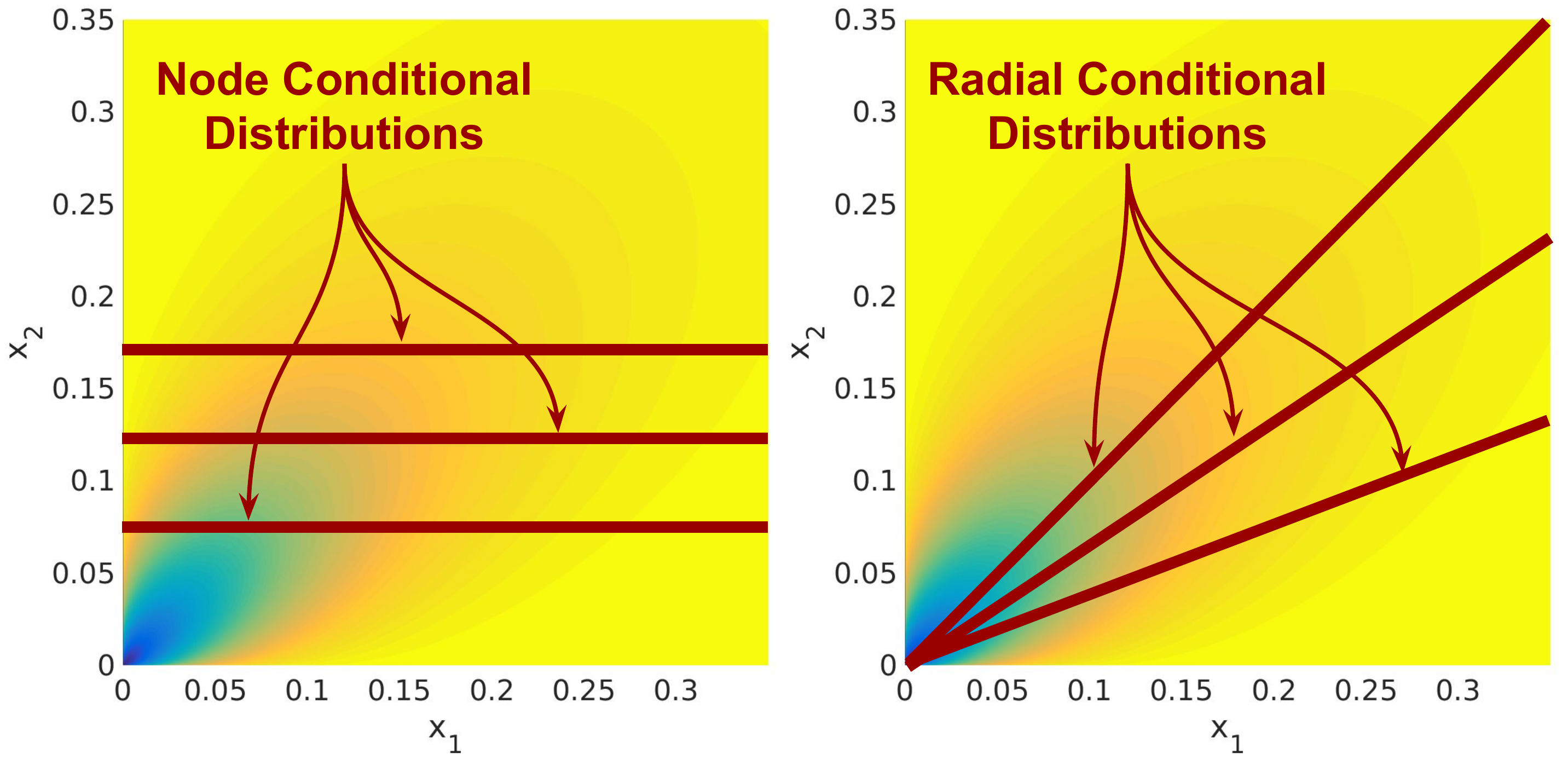}
\vspace{-1.8em}
\caption{\emph{Node} conditional distributions (left) are univariate probability distributions of one variable assuming the other variables are given while \emph{radial} conditional distributions are univariate probability distributions of vector scaling assuming the vector direction is given.  Both conditional distributions are helpful in understanding SQR graphical models.}
\label{fig:conditional-illustration}
\end{figure}

We analyze two types of univariate conditional distributions of the SQR graphical models.  The first is the standard \emph{node} conditional distribution, i.e. the conditional distribution of one variable given the values for all other variables (see Fig.~\ref{fig:conditional-illustration}).  The second is what we will call the \emph{radial} conditional distribution in which the \emph{unit direction} is fixed but the length of the vector is unknown (see Fig.~\ref{fig:conditional-illustration}).  The node conditional distribution is helpful for parameter estimation as described more fully in Sec.~\ref{sec:parameter-estimation}.  The radial conditional distribution is important for understanding the form of the SQR distribution as well as providing a means to succinctly prove that the normalization constant is finite (i.e. the distribution is valid) as described in Sec.~\ref{sec:normalization}.

\begin{figure}[!ht]
\centering
\includegraphics[width=0.49\columnwidth, trim = 7.05cm 10.1cm 7.5cm 10.3cm, clip]{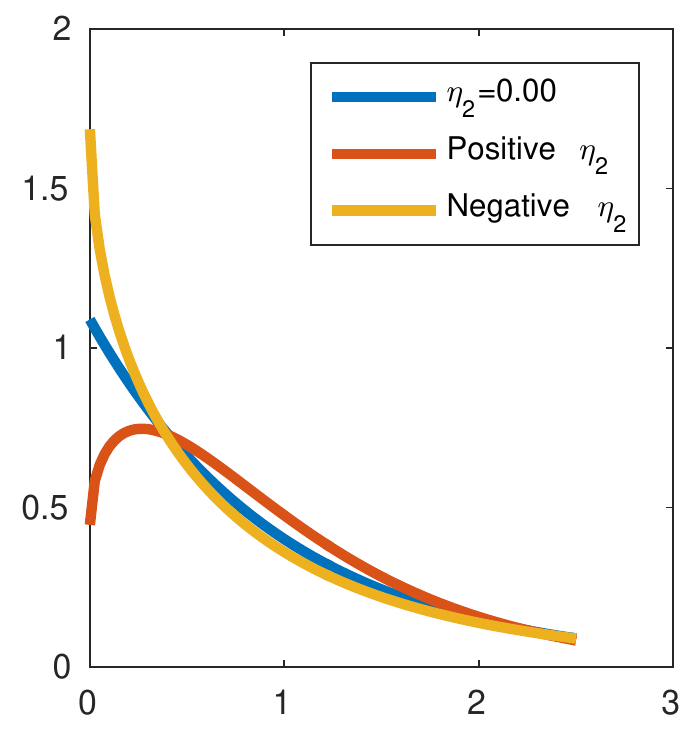}
\includegraphics[width=0.49\columnwidth, trim = 7.05cm 10.1cm 7.5cm 10.3cm, clip]{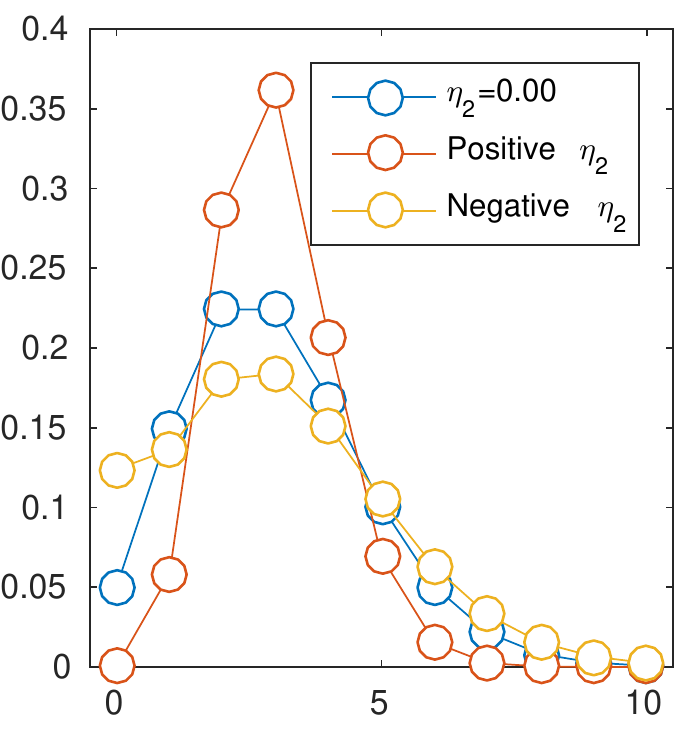}
\caption{Examples of the node conditional distributions of exponential (left) and Poisson (right) SQR models for $\eta_2 = 0$, $\eta_2 > 0$ and $\eta_2 < 0$.}
\label{fig:node-conditionals}
\end{figure}

\paragraph{Node Conditional Distribution}
The probability distribution of one variable $\inst_\vi$ given all other variables $\instvec_{-\vi} = [\inst_1,\inst_2,\dots,\inst_{\vi-1},\inst_{\vi+1},\dots,\inst_\vmax ]$ is as follows:
\begin{align*}
&\begin{array}{l}
\Pr(\inst_\vi \,|\, \instvec_{-\vi}, \nodepvec, \pmat) \propto \\
\exp\!\Big\{\!\psca_{\vi\vi}\T(\inst_\vi) + \Big(\!\nodep_\vi\! \!+\! 2 \pvec_{-\vi}^T \!\sqrt{\T(\instvec_{-\vi})}\Big) \!\sqrt{\T(\inst_\vi)} \!+\! \B(\inst_\vi) \! \Big\},
\end{array}
\end{align*}
where $\pvec_{-\vi} \in \R^{\vmax-1}$ is the $\vi$-th column of $\pmat$ with the $\vi$-th entry removed.  This conditional distribution can be reformulated as a new two parameter exponential family:
\begin{align}
&\begin{array}{l}
\Pr(\inst_\vi \,|\, \instvec_{-\vi}, \nodepvec, \pmat) = \\
\exp\Big( \natp_1 \tilde{\T}_1(\inst_\vi) +   \natp_2 \tilde{\T}_2(\inst_\vi) + \B(\inst_\vi) - \A_{\text{node}}(\natpvec) \Big)
\end{array} \label{eqn:node-conditional} 
\end{align}
\begin{align}
&\begin{array}{l}
\A_{\text{node}}(\natpvec) = \\
\int_{\domain}  \exp\Big( \natp_1 \tilde{\T}_1(\inst_\vi) +   \natp_2 \tilde{\T}_2(\inst_\vi) + \B(\inst_\vi) \Big) \mathrm{d}\measure(\inst_\vi) \, ,
\end{array} \label{eqn:node-conditionalA}
\end{align}
where $\natp_1 = \psca_{\vi\vi}$, $\natp_2 = \nodep_\vi + 2 \pvec_{-\vi}^T \sqrt{\T(\instvec_{-\vi})} $, $\tilde{\T}_1(\inst) =\T(\inst)$, and $\tilde{\T}_2(\inst) = \sqrt{\T(\inst)}$. Note that this reduces to the base exponential family only if $\natp_2 = 0$ unlike the model in Eqn.~\ref{eqn:glm-gm} which, by construction, has node conditionals in the base exponential family.  Examples of node conditional distributions for the exponential and Poisson SQR can be seen in Fig.~\ref{fig:node-conditionals}.  While these node conditionals are different from the base exponential family and hence slightly more difficult to use for parameter estimation as described later in Sec.~\ref{sec:parameter-estimation}, the benefit of almost arbitrary positive and negative dependencies significantly outweighs the cost of using SQR over previous graphical models.

\paragraph{Radial Conditional Distribution}
For simplicity, let us assume w.l.o.g. that $\T(\instvec) = \instvec$.\footnote{If $\T$ is not linear than we can merely reparameterize the distribution so that this is the case.}  Suppose we condition on the unit direction $\uvec = \frac{\instvec}{\|\instvec\|_1}$ of the sufficient statistics but the scaling of this unit direction $\radx = \|\instvec\|_1$ is unknown.  We call this the \emph{radial} conditional distribution:
\begin{align*}
\begin{array}{l}
\Pr(\instvec = \radx \uvec \,|\, \uvec, \nodepvec, \pmat) \\
\propto \exp\Big( \nodepvec^T \sqrt{\radx \uvec} + \sqrt{\radx \uvec}^T \,\pmat\, \sqrt{\radx \uvec} + \sum_\vi \B(\radx\usca_\vi) \Big) \\
\propto \exp\Big( (\nodepvec^T \sqrt{ \uvec}) \sqrt{\radx} + \big(\sqrt{ \uvec}^T \,\pmat\, \sqrt{\uvec}\big) \radx + \sum_\vi \B(\radx\usca_\vi) \Big)\, .
\end{array}
\end{align*}
The radial conditional distribution can be rewritten as a univariate exponential family:
\begin{align}
&\Pr(\radx \,|\, \uvec, \nodepvec, \pmat) = \exp\Big( \underbrace{\bar{\natp}_1 \radx + \bar{\natp}_2 \sqrt{\radx}}_{O(\radx)}  + \underbrace{\tilde{\B}_{\uvec}(\radx)}_{O(\B(\radx))} - \A_{\text{rad}}(\bar{\natpvec}) \Big) \label{eqn:radial-conditional} \\
&\A_{\text{rad}}(\bar{\natpvec}) = \int_{\domain} \exp\Big( \underbrace{\bar{\natp}_1 \radx + \bar{\natp}_2 \sqrt{\radx}}_{O(\radx)}  + \underbrace{\tilde{\B}_{\uvec}(\radx)}_{O(\B(\radx))} \Big) \mathrm{d}\measure(\radx) \, , \label{eqn:radial-conditionalA}
\end{align}
where $\bar{\natp}_1 = \sqrt{\uvec}^T \pmat \sqrt{\uvec}$, $\, \bar{\natp}_2 = \nodepvec^T \sqrt{\uvec}$ and $\tilde{\B}_{\uvec}(\radx) = \sum_\vi \B(\radx\usca_\vi)$.  Note that if the log base measure of the base exponential family is zero $\B(\inst) = 0$, then the radial conditional is the same as the node conditional distribution because the modified base measure is also zero $\tilde{\B}_{\uvec}(\radx) = 0$.  If both $\nodepvec = 0$ and $\B(\inst)=0$, this actually reduces to the base exponential family. For example, the exponential distribution has $\B(\inst) = 0$, and thus if we set $\nodepvec = 0$, the radial conditional of an exponential SQR is merely the exponential distribution.  Other examples with a log base measure of zero include the Beta distribution and the gamma distribution with a known shape.  For distributions in which the log base measure is not zero, the distribution will deviate from the node conditional distribution based on the relative difference between $\B(\inst)$ and $\tilde{\B}_{\uvec}(\inst)$.  However, the important point even for distributions with non-zero log base measures is that the terms in the exponent grow at the same rate as the base exponential family---i.e. $O(\radx) + O(\B(\radx))$.  This helps to ensure that the radial conditional distribution is normalizable even as $\radx \to \infty$ since the base exponential family was normalizable.  As an example, the Poisson distribution has the log base measure $\B(\inst) = -\log(\inst!)$ and thus $\tilde{\B}_{\uvec}(\inst)$ is $O(-\inst \log \inst)$ whereas the other terms $\bar{\natp}_1\radx + \bar{\natp}_2\sqrt{\radx}$ are only $O(\radx)$.  This provides the intuition of why the Poisson SQR radial distribution is normalizable as will be explained in Sec.~\ref{sec:poisson-model}.

\subsection{Normalization}
\label{sec:normalization}
Normalization of the distribution was the reason for the negative-only parameter restrictions of the exponential and Poisson distributions in the previous graphical models \citep{Besag1974,Yang2015} as defined in Eqn.~\ref{eqn:glm-gm}.  However, we show that in the case of SQR models, normalization is much simpler to achieve and generally puts little to no restriction on the value of the parameters---thus allowing both positive and negative dependencies.  For our derivations, let $\Uset = \{ \uvec : \|\uvec\|_1 = 1, \uvec \in \Rp^\vmax\}$ be the set of unit vectors in the positive orthant. The SQR log partition function $\A(\pmat)$ can be decomposed into nested integrals over the unit direction and the one dimensional integral over scaling, denoted $\radx$:
\begin{align}
&\A(\nodepvec,\pmat) = \log \!\! \int\limits_\Uset \!\! \int\limits_{\radset(\uvec)} \!\!\!\! \exp\Big( \nodepvec^T\sqrt{\radx \uvec}  +  \sqrt{\radx \uvec}^T \pmat \sqrt{\radx \uvec} \\
&\hspace{7em} + \sum_\vi \B(\radx\usca_\vi) \Big) \mathrm{d}\measure(\radx) \, \mathrm{d}\uvec \notag \\
&= \log \!\! \int\limits_\Uset \!\! \int\limits_{\radset(\uvec)} \!\!\!\!
\exp(\bar{\natp}_1(\uvec) \radx \!+\! \bar{\natp}_2(\uvec) \sqrt{\radx} \!+\! \sum_\vi \B(\radx\usca_\vi) ) \mathrm{d}\measure(\radx) \, \mathrm{d}\uvec ,
\end{align}
where $\radset(\uvec) = {\{\radx \in \Rp:\radx\uvec \in \domain\}}$, and $\measure$ and $\domain$ are defined as in Eqn.~\ref{eqn:glm-gmA}.  Because $\Uset$ is bounded, we merely need that the radial conditional distribution is normalizable (i.e. $\A_{\text{rad}}(\bar{\natpvec}) < \infty$ from Eqn.~\ref{eqn:radial-conditionalA}) for the joint distribution to be normalizable. As suggested in Sec.~\ref{sec:conditionals}, the radial conditional distribution is similar to the base exponential family and thus likely only has similar restrictions on parameter values as the base exponential family.  In Sec.~\ref{sec:gaussian-sqr}, we give examples for the exponential SQR and Poisson SQR distributions showing that this condition can be achieved with little or no restriction on the parameter values.

\subsection{Parameter Estimation}
\label{sec:parameter-estimation}
For estimating the parameters $\pmat$ and $\nodepvec$, we follow the basic approach of \citep{Ravikumar2010,Yang2015,Yang2013} and fit $\vmax$ $\ell_1$-regularized node-wise regressions using the node conditional distributions described in Sec.~\ref{sec:conditionals}. Thus, given a data matrix $\instmat \in \R^{\vmax \times \imax}$ we attempt to optimize the following convex function:
\begin{align}
\begin{array}{l}
\argmin\limits_{\pmat} -\frac{1}{\imax}\sum_\vi \sum_\ii \Big( \natp_{1\vi\ii}\inst_{\vi\ii} + \natp_{2\vi\ii} \sqrt{\inst_{\vi\ii}} \\
\hspace{3em} + \B(\inst_{\vi\ii}) - \A_{\text{node}}(\natp_{1\vi\ii},\natp_{2\vi\ii}) \Big) + \lambda \|\pmat\|_{1,\text{off}} \, ,
\end{array}
\label{eqn:optimization}
\end{align}
where $\natp_{1\vi\ii} = \psca_{\vi,\vi}$, $\natp_{2\vi\ii} = \nodep + 2 \pvec_{-\vi}^T \sqrt{\T(\instvec_{-\vi\ii})}$, $\|\pmat\|_{1,\text{off}} = \sum_{\vi\neq\vit} |\psca_{\vi\vit}|$ is the $\ell_1$-norm on the off diagonal elements and $\lambda$ is a regularization parameter.  Note that this can be trivially parallelized into $\vmax$ independent sub problems which allows for significantly faster computation as in \citep{Inouye2015}. Unlike previous graphical models \citep{Yang2015} that were known to have closed-form solutions to the node conditional log partition function, the main difficulty for SQR graphical models is that the node conditional log partition function $\A_{\text{node}}(\natpvec)$ is not known to have a closed form in general.

For the particular case of exponential SQR models, there is a closed-form solution for $\A_{\text{node}}$ using the error function as will be seen in Sec.~\ref{sec:exponential-model} on exponential SQR models.  More generally, because $\A_{\text{node}}$ is merely a one dimensional summation or integral, standard numerical approximations such as Gaussian quadrature could be used.  Similarly, the gradient of $\nabla \A_{\text{node}}$ could be numerically approximated by:
\begin{align}
\begin{array}{l}
\nabla \A_{\text{node}} = \frac{1}{\epsilon}\Big[ \big(\hat{\A}(\natp_1+\epsilon,\natp_2) - \hat{\A}(\natp_1,\natp_2)\big), \\ 
\hspace{6em}\big(\hat{\A}(\natp_1,\natp_2+\epsilon) - \hat{\A}(\natp_1,\natp_2)\big) \Big] \, ,
\end{array}
\end{align}
where $\epsilon$ is a small step such as 0.001.  Notice that to compute the function value and the gradient, only three  1D numerical integrations are needed.  Another significant speedup that could be explored in future work would be to use a Newton-like method as in \citep{Hsieh2014,Inouye2015}, which optimize a quadratic approximation around the current iterate.  Because these Newton-like methods only need a small number of Newton iterations to converge, the number of numerical integrations could be reduced significantly compared to gradient descent which often require thousands of iterations to converge.

\subsection{Likelihood Approximation}
\label{sec:likelihood}
We use Annealed Importance Sampling (AIS) \citep{Neal2001} similar to the sampling used in \citep{Inouye2015} for likelihood approximation.  In particular, we need to approximate the SQR log partition function $\A(\nodepvec,\pmat)$ as in Eqn.~\ref{eqn:sqr-modelA}.  First, we derive a slice sample for the node conditionals in which the bounds for the slice can be computed in closed form. Second, we use the slice sampler to develop a Gibbs sampler for SQR models.  Finally, we derive an annealed importance sampler \citep{Neal2001} using the Gibbs sampler as the intermediate sampler by linearly combining the off-diagonal part of the parameter matrix $\pmat_{\text{off}}$ with the diagonal part $\pmat_{\text{diag}}$---i.e. $\tilde{\pmat} = \aisscale\pmat_{\text{off}} + \pmat_{\text{diag}}$. We also modify $\tilde{\nodepvec} = \aisscale\nodepvec$ similarly. For each successive distribution, we linearly change $\aisscale$ from 0 to 1.  Thus, we start by sampling from the base exponential family independent distribution $\prod_{\vi=1}^\vmax \Pr(\instvec \,|\, \natp_{1\vi} = \psca_{\vi\vi}, \natp_{2\vi} = 0)$ and slowly move towards the final SQR distribution $\Pr(\instvec \,|\, \nodepvec, \pmat)$.  We maintain the sample weights as defined in \citep{Neal2001} and from these weights, we can compute an approximation to the log partition function \citep{Neal2001}.

\section{Examples from Various Exponential Families}
\label{sec:gaussian-sqr}
We give several examples of SQR graphical models in the following sections (however, it should be noted that we have been developing a \emph{class} of graphical models for \emph{any} univariate exponential family with nonnegative sufficient statistics). The main analysis for each case is determining what conditions on the parameter matrix $\pmat$ allow the joint distribution to be normalized. As described in Sec.~\ref{sec:normalization}, for SQR models, this merely reduces to determining when the radial conditional distribution is normalizable.  We analyze the exponential and Poisson cases in later sections but first we give examples of the discrete and Gaussian SQR graphical models.

The discrete SQR graphical model---including the binary Ising model---is equivalent to the standard discrete graphical model because the sufficient statistics are indicator functions $\T_\vi(\inst) = \bm{I}(\inst = \vi), \forall \vi\neq\vmax$ and the square root of an indicator function is merely the indicator function.  Thus, in the discrete case, the discrete graphical model in \citep{Ravikumar2010,Yang2015} is equivalent to the discrete SQR graphical model. For the Gaussian distribution, we can use the nonnegative Gaussian sufficient statistic $\T(\inst) = \inst^2$.  Thus, the Gaussian SQR graphical model is merely $\Pr(\instvec | \pmat) \propto \exp(\nodepvec^T \instvec + \instvec^T\pmat\instvec)$, which by inspection is clearly the standard Gaussian distribution where $\nodepvec = \Sigma^{-1}\mu$ and $\pmat = -\frac{1}{2}\Sigma^{-1}$ is required to be negative definite.\footnote{This is by the slightly nuanced definition of the square root operator in Eqn.~\ref{eqn:sqr-model} and \ref{eqn:sqr-modelA} such that $\sqrt{\inst^2} \equiv \inst$ rather than $|\inst|$.}  Thus, the Gaussian graphical model can be seen as a special case of SQR graphical models.

\subsection{Exponential SQR Graphical Model}
\label{sec:exponential-model}
We consider what are the required conditions on the parameters $\nodepvec$ and $\pmat$ for the exponential SQR graphical model. If $\bar{\natp}_1$ is positive, the log partition function will diverge because even the end point $\lim_{\radx \to \infty} \exp(\bar{\natp}_1 \radx) \to \infty$.  On the other hand, if $\bar{\natp}_1$ is negative, then the radial conditional distribution is similar in form to the exponential distribution and thus the log partition function will be finite because the negative linear term $\bar{\natp}_1 \radx$ dominates in the exponent as $\radx \to \infty$.\footnote{On the edge case when $\bar{\natp}_1 = 0$, the log partition function will diverge if $\bar{\natp}_2 \geq 0$ and will converge if $\bar{\natp}_1 < 0$ by simple arguments. The normalizability condition when $\natp_2 = 0$ could slightly loosen the condition on $\pmat$ in Eqn.~\ref{eqn:exp-condition} but for simplicity we did not include this edge case.} See appendix for proof.  Thus, the basic condition on $\Phi$ is:
\begin{align}
\Phi_{\text{Exp}} \in \{ \Phi : \sqrt{\uvec}^T\pmat \sqrt{\uvec} < 0, \forall \uvec \in \Uset \} \,. \label{eqn:exp-condition}
\end{align}
Note that this allows both positive and negative dependencies.  A sufficient  condition is that $\Phi$ be negative definite---as is the case for Gaussian graphical models.  However, negative definiteness is far from necessary because we only need negativity of the interaction term for vectors in the positive orthant.  It may even be possible for $\pmat$ to positive definite but Eqn.~\ref{eqn:exp-condition} be satisfied; however, we have not explored this idea.

For fitting the SQR model, the node conditional log partition function $\A_{\text{Exp}}(\natpvec)$ has a closed-form solution:
\begin{align*}
\A_{\text{Exp}}(\natpvec) &= \!\log\!\bigg(\! \frac{\sqrt{\pi}\natp_1\exp\big(\frac{{-\natp}_2^2}{4\natp_1}\big)\!\Big(\! 1 \!-\! \operatorname{erf}\!\big(\frac{-\natp_2}{2\sqrt{-\natp_1}}\big)\!\Big) }{{-2}({-\natp_1})^{\frac{3}{2}}} \!-\! \frac{1}{\natp_1} \!\bigg) ,
\end{align*}
where $\operatorname{erf}(\cdot)$ is the error function.  The $\operatorname{erf}$ function shows up because of an initial substitution of $u = \sqrt{x}$ to transform the exponent into a quadratic form.  Note that $\natp_1 < 0$ by the condition on $\pmat_{\text{Exp}}$ in Eqn.~\ref{eqn:exp-condition} above.  The derivatives of $\A_{\text{Exp}}$ can also be computed in closed form for use in the parameter estimation algorithm.

\subsection{Poisson SQR Graphical Model}
\label{sec:poisson-model}
The normalization analysis for Poisson SQR graphical model is also relatively simple but requires a more careful analysis than the exponential SQR graphical model.  Let us consider the form of the Poisson radial conditional:
$\Pr_{\text{rad}}(\radx \,|\, \uvec ) \propto \exp( \bar{\natp}_1\radx + \bar{\natp}_2 \sqrt{\radx} - \sum_\vi \log((\radx\usca_\vi)!)).$  Note that the domain of $\radx$, denoted $\domain_\radx = \{\radx \in \Zp: \radx\uvec \in \Zp^\vmax \}$, is discrete. We can simplify the analysis by taking a larger domain $\tilde{\domain}_\radx = \{\radx \in \Zp\}$ of all non-negative integers and changing the log factorial to the smooth gamma function, i.e. $\sum_\vi \log((\radx\usca_\vi)!)) \to \sum_\vi \log(\Gamma(\radx\usca_\vi+1))$.  Thus, the radial conditional log partition function is upper bounded by:
\begin{align}
\sum_{\radx \in \Zp} \!\!\exp\!\Big(\! \underbrace{\bar{\natp}_1\radx + \bar{\natp}_2 \sqrt{\radx}}_{O(\radx)} - \underbrace{{\textstyle \sum_\vi} \log(\Gamma(\radx\usca_\vi+1))}_{O(\radx \log \radx)} \!\Big) \!<\! \infty. \label{eqn:poisson-norm}
\end{align}
The basic intuition is that the exponent has a linear $O(\radx)$ term minus an $O(\radx \log \radx)$ term, which will eventually overcome the linear term and hence the summation will converge. Note that we did not assume any restrictions on $\pmat$ except that all the entries are finite.  Thus, for the Poisson distribution, $\pmat$ can have arbitrary positive and negative dependencies.  A formal proof for Eqn.~\ref{eqn:poisson-norm} is given in the appendix.

\section{Experiments and Results}
\subsection{Synthetic Experiment}
\label{sec:synthetic-exp}
In order to show that our parameter estimation algorithm has the ability to find the correct dependencies, we develop a synthetic experiment on chain-like graphs.  We construct $\Phi$ to be a $k$-dependent circular chain-like graph by first setting the diagonal of $\Phi$ to be 1.  Then, we add an edge between each node and its $k$ neighbors with a value of $\frac{0.9}{2*k}$, i.e. the $\vi$-th node is connected to the $(\vi+1)$-th, $(\vi+2)$-th, $\dots$, $(\vi + k)$-th nodes where the indices are modulo $\vmax$ (e.g. $k=1$ is the standard chain graph).  This ensures that $\Phi$ is negative definite by the Gershgorin disc theorem.  We generate samples using Gibbs sampling with 1000 Gibbs iterations per sample and 10 slice samples for each node conditional sample.  For this experiment, we set $\vmax = 30, \lambda = 10^{-5}, k \in \{ 1,2,3,4\}$, and $\imax \in \{100,200,400,800,1600\}$.  We calculate the edge precision for the fitted model by computing the precision for the top $k\vmax$ edges---i.e. the number of true edges in the top $k\vmax$ estimated edges over the total number of true edges.  The results in Fig.~\ref{fig:airport-results} demonstrate that our parameter estimation algorithm is able to easily find the edges for small $k$ and is even able to identify the edges for large $k$, though the problem becomes more difficult when $k$ is large (because there are more parameters, which are also smaller), and thus more samples are needed.  With 1,600 samples, our parameter estimation algorithm is able to recover at least 95\% of the edges even when $k = 4$.

\subsection{Airport Delay Times Experiment}
In order to demonstrate that the SQR graphical model class is more suitable for real-world data than the graphical models in \citep{Yang2015} (which can only model \emph{negative} dependencies), we fit an exponential SQR model to a dataset of airport delay times at the top 30 commercial USA airports---also known as Large Hub airports. We gathered flight data from the US Department of Transportation public ``On-Time: On-Time Performance'' database\footnote{http://www.transtats.bts.gov/DL\_SelectFields.asp?Table\_ID =236\&DB\_Short\_Name=On-Time} for the year 2014.  We calculated the average delay time per day at each of the top 30 airports (excluding cancellations).

For our implementation, we set $\lambda \in \{0.05,0.005,0.0005\}$ and set a maximum of 5000 iterations for our proximal gradient descent algorithm.  
For approximating the log partition function using the AIS sampling defined in Sec.~\ref{sec:likelihood}, we sampled 1000 AIS samples with 100 annealing distributions---i.e. $\aisscale$ took 100 values between 0 and 1---, 10 Gibbs steps per annealed distribution and 10 slice samples for every node conditional sampling.  Generally, our algorithm with these parameter settings took roughly 35 seconds to train the model and about 25 seconds to compute the likelihood (i.e. AIS sampling) using MATLAB prototype code on the TACC Maverick cluster.\footnote{https://portal.tacc.utexas.edu/user-guides/maverick}

\begin{figure}[!ht]
\centering
\includegraphics[width=0.39\columnwidth, trim = 6cm 2.5cm 5.5cm 2.2cm, clip]{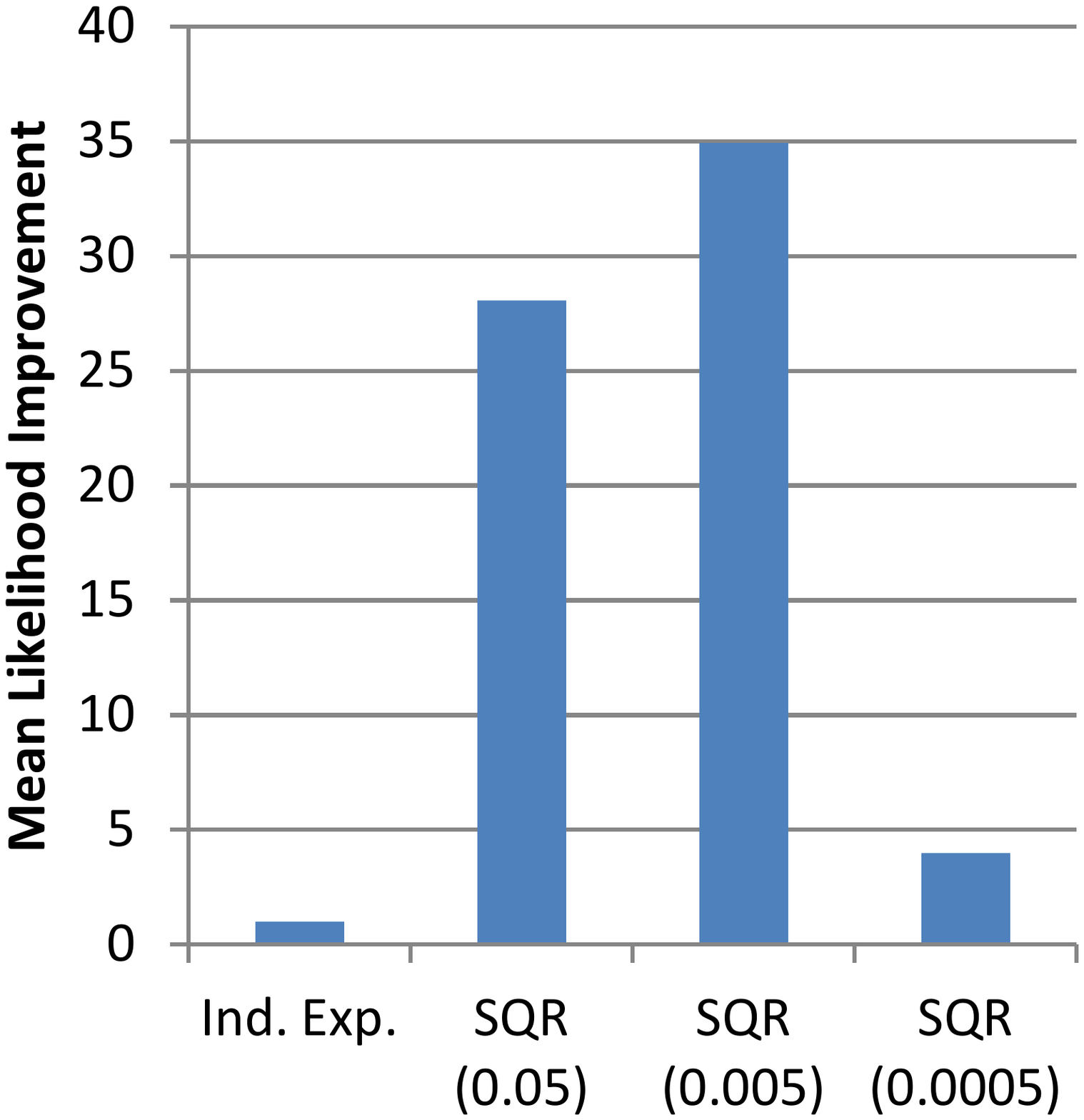}
\includegraphics[width=0.59\columnwidth, trim = 2.5cm 3.1cm 2.5cm 2.5cm, clip]{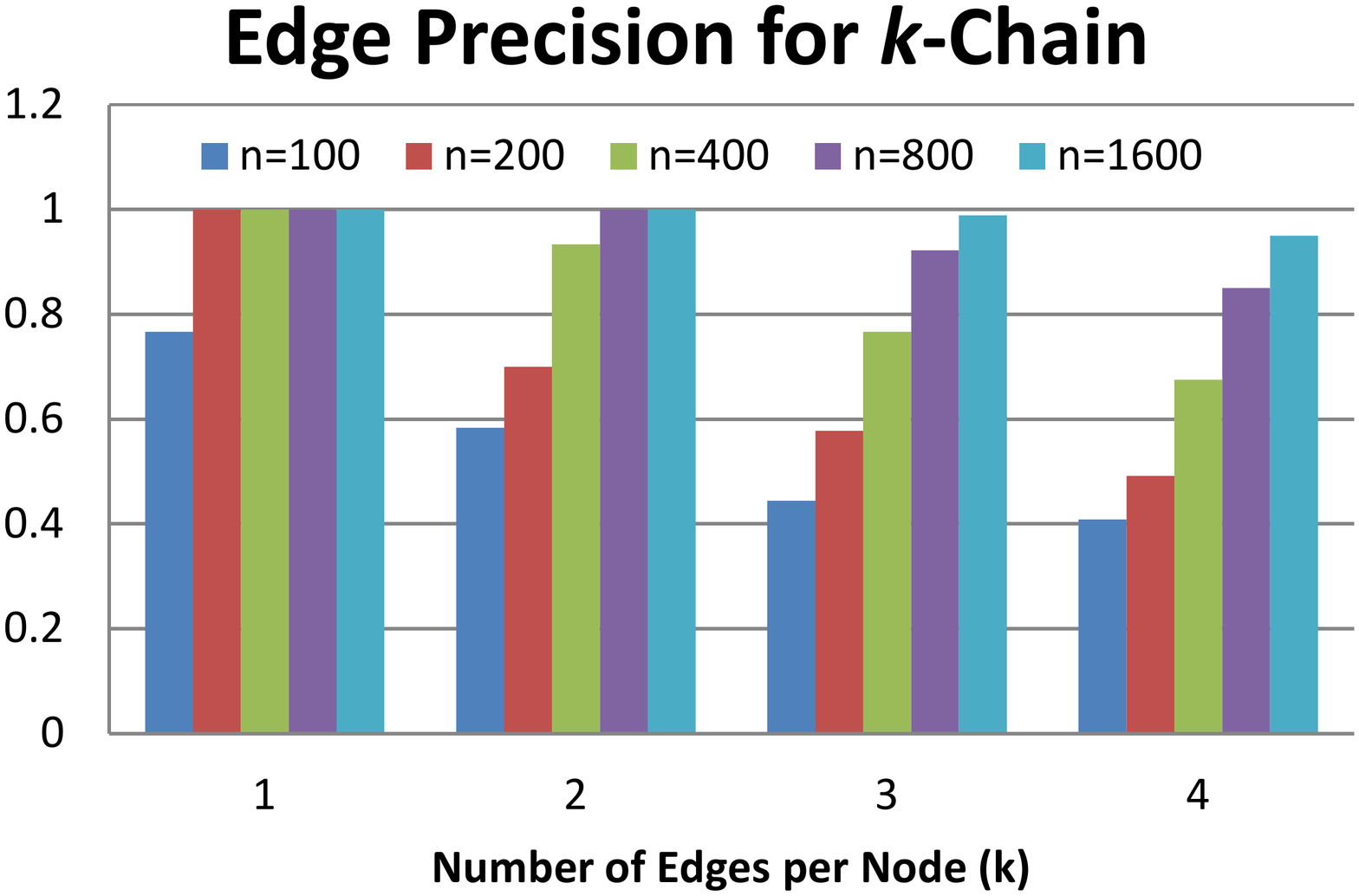}
\vspace{-1em}
\caption{(Left) The fitted exponential SQR model improves significantly over the independent exponential model in terms of relative likelihood suggesting that a model with positive dependencies is more appropriate. (Right) The edge precision for the circular chain graph described in Sec.~\ref{sec:synthetic-exp} demonstrate that our parameter estimation algorithm is able to effectively identify edges for small $k$, and if given enough samples, can also identify edges for larger $k$.}
\label{fig:airport-results}
\end{figure}

\begin{figure}[!ht]
\centering
\includegraphics[width=\columnwidth]{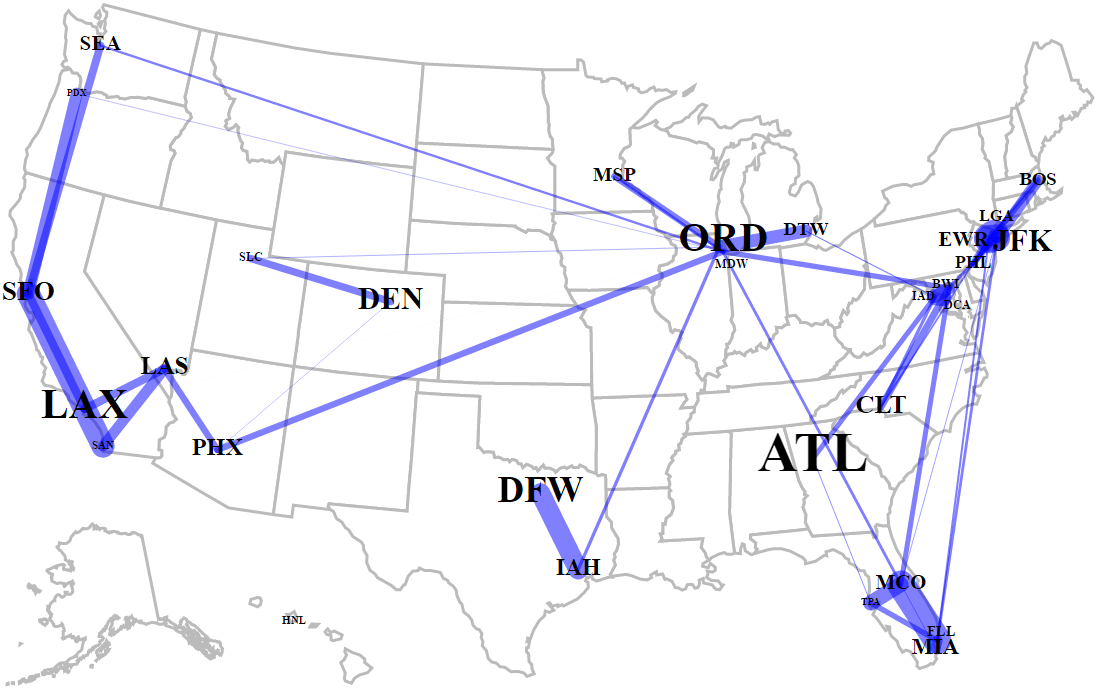}
\caption{Visualizing the top 50 edges between airports shows that SQR models can capture interesting and intuitive \emph{positive} dependencies even though previous exponential graphical models \cite{Yang2015} were restricted to negative dependencies.  The delays at the Chicago airports seem to affect other airports as would be expected because of Chicago weather delays.  Other dependencies are likely related to weather or geography.  (For this visualization, we set  $\lambda = 0.0005$.  Width of lines is proportional to the value of the edge weight, i.e. a non-zero in $\Phi$, and the size of airport abbreviation is proportional to the average number of passengers.) }
\label{fig:airport-visualization}
\end{figure}

We computed the geometric mean of the relative log likelihood compared to the independent exponential model, i.e. $\exp(( \mathcal{L}_{\text{SQR}} - \mathcal{L}_{\text{Ind}} )/\imax)$, where $\mathcal{L}$ is the log likelihood.  These values can be seen in Fig.~\ref{fig:airport-results} (higher is better). Clearly, the exponential SQR model provides a major improvement in relative likelihood over the independent model suggesting that the delay times of airports are clearly related to one another.  In Fig.~\ref{fig:airport-visualization}, we visualize the non-zeros of $\Phi$---which correspond to the edges in the graphical model---to show that our model is capturing intuitive positive dependencies.

First, it should be noted that all the dependencies are positive yet positive dependencies were not allowed by previous graphical models \citep{Yang2015}!  Second, as would be expected because of weather delays, the airports in the Chicago area seems to affect the delays of many other airports.  Similarly, a weather effect seems to be evident for the airports near New York City.   Third, as would be expected, some dependencies seem to be geographic in nature as seen by the west coast dependencies, Texas dependency (i.e. DFW-IAH), and east coast dependencies.  Note that the geographic dependencies were found even though no location data was given to the algorithm.  Fourth, the busiest airport in Atlanta, GA (ATL) is not strongly dependent on other airports.  This seems reasonable because Atlanta rarely has snow and there are few major airports geographically close to Atlanta.  These qualitative results suggest that the exponential SQR model is able to capture multiple interesting and intuitive dependencies.

\section{Discussion}
As full probability models, SQR graphical models could be used in any situation where a multivariate distribution is required.  For example, SQR models could be used in Bayesian classification by modeling the probability of each class distribution instead of the classical Naive Bayes assumption of independence.  As another example, SQR models could be used as the base distribution in mixtures or admixture composite distributions as in \citep{Inouye2014,Inouye2014b}---similar to multivariate Gaussian mixture models.  Another extension would be to consider mixed SQR graphical models in which the joint distribution has variables using different exponential families as base distributions as explored for previous graphical models in \citep{Yang2014c,Tansey2015}.

\section{Conclusion}
We introduce a novel class of graphical models that creates multivariate generalizations for \emph{any} univariate exponential family with nonnegative sufficient statistics---including Gaussian, discrete, exponential and Poisson distributions.  We show that SQR graphical models generally have few restrictions on the parameters and thus can model \emph{both positive and negative} dependencies unlike previous generalized graphical models as represented by \citep{Yang2015}.  In particular, for the exponential SQR model, the parameter matrix $\pmat$ can have both positive and negative dependencies and is only constrained by a mild condition---akin to the positive-definiteness condition on Gaussian covariance matrices.  For the Poisson distribution, there are no restrictions on the parameter values, and thus the Poisson SQR model allows for arbitrary positive and negative dependencies.  We develop parameter estimation and likelihood approximation methods and demonstrate that the SQR model indeed captures interesting and intuitive dependencies by modeling both synthetic datasets and a real-world dataset of airport delays. The general SQR class of distributions opens the way for graphical models to be effectively used with non-Gaussian and non-discrete data without the unintuitive restriction to negative dependencies.

\clearpage
\section*{Acknowledgements} 
This work was supported by NSF (DGE-1110007, IIS-1149803, IIS-1447574, IIS-1546459, DMS-1264033, CCF-1320746) and ARO (W911NF-12-1-0390).

\bibliography{static-all}
\bibliographystyle{icml2016}

\stop
\clearpage
\appendix
\section{Proof of Exponential SQR Normalization}
The basic intuition is clear by looking at the asymptotic growth of each term.  However, we specifically outline the possibilities:
\begin{enumerate}
\item $\natp_2 > 0$: $\A(\natp_1,\natp_2) \to \infty$.
\item $\natp_2 < 0$: $\A(\natp_1,\natp_2) < \infty$.
\item $\natp_2 = 0$: if $\natp_1 < 0$, then $\A(\natp_1,\natp_2) < \infty$, otherwise $\A(\natp_1, \natp_2) \to \infty$.
\end{enumerate}
In summary, we need that $\natp_2 < 0$ \emph{or} ($\natp_2 = 0$ \emph{and} $\natp_1 < 0$).

\paragraph{Case 1: $\natp_2 > 0$} Let $\hat{\natp}_2 = \natp_2/2$.  First, we seek an exponential lower bound on the partition function.  In particular, we want to find an $\bar{\radx}$ such that for all $\radx > \bar{\radx}$, $\exp(\hat{\natp}_2 \radx) \leq \exp(\natp_1 \sqrt{\radx} + \natp_2 \radx)$.  Taking the log of both sides and solving, we find that the critical points of the above inequality are at 0 and $(-2\frac{\natp_1}{\natp_2})^2$.  We take the non-trivial solution of $\bar{\radx} = (-2\frac{\natp_1}{\natp_2})^2$.  Now we need to check if the region to the right of $\bar{\radx}$ is possible by plugging into the original equation.  Let us try a point $\tilde{\radx} = a \bar{\radx}$ where $a > 1$:
\begin{align}
\exp(\hat{\natp}_2 \tilde{\radx}) &\stackrel{?}{\leq} \exp(\natp_1 \sqrt{\tilde{\radx}} + \natp_2 \tilde{\radx}) \\
\Rightarrow \hat{\natp}_2 \tilde{\radx} &\stackrel{?}{\leq} \natp_1 \sqrt{\tilde{\radx}} + \natp_2 \tilde{\radx} \\
\Rightarrow (\natp_2/2 - \natp_2) \tilde{\radx} &\stackrel{?}{\leq} \natp_1 \sqrt{\tilde{\radx}} \\
\Rightarrow -\frac{\natp_2}{2} \left(-2a\frac{\natp_1}{\natp_2}\right)^2 &\stackrel{?}{\leq} \natp_1 \sqrt{\left(-2a\frac{\natp_1}{\natp_2}\right)^2} \\
\Rightarrow a\frac{-2\natp_1^2}{\natp_2} &\stackrel{?}{\leq}  \sqrt{a} \frac{-2\natp_1^2}{\natp_2} \label{eqn:before-multiply} \\
\Rightarrow a &\geq  \sqrt{a} \, ,\label{eqn:lower-bound}
\end{align}
where the last line is because we assumed $a > 1$ and $\natp_2 > 0$.  Thus, we can lower bound the log partition function as follows:
\begin{align*}
\A(\natp_1,\natp_2) &= \int_0^{\bar{\radx}} \exp(\natp_1\sqrt{\radx}+\natp_2\radx) \mathrm{d}\radx \\
&\hspace{5em} + \int_{\bar{\radx}}^\infty \exp(\natp_1\sqrt{\radx}+\natp_2\radx) \mathrm{d}\radx \\
&\geq \int_0^{\bar{\radx}} \exp(\natp_1\sqrt{\radx}+\natp_2\radx) \mathrm{d}\radx + \underbrace{\int_{\bar{\radx}}^\infty \exp(\hat{\natp}_2\radx) \mathrm{d}\radx}_{\to \infty, \,\text{since}\, \hat{\natp}_2 > 0} \\
&= \infty \,.
\end{align*}
Therefore, if $\natp_2 > 0$, the log partition function diverges and hence the joint distribution is not consistent.

\paragraph{Case 2: $\natp_2 < 0$} Now we will find an exponential upper bound and show that this upper bound converges---and hence the log partition function converges. In a similar manner to case 1, let $\hat{\natp}_2 = \natp_2/2$.  We want to find an $\bar{\radx}$ such that for all $\radx > \bar{\radx}$, $\exp(\hat{\natp}_2 \radx) \geq \exp(\natp_1 \sqrt{\radx} + \natp_2 \radx)$---the only difference from case 1 is the direction of the inequality.  Thus, using the same reasoning as case 1, we have that $\bar{\radx} = (-2\frac{\natp_1}{\natp_2})^2$.  Similarly, we need to check if the region to the right of $\bar{\radx}$ is possible by plugging into the original equation.  In a analogous derivation, we arrive at the same equation as Eqn.~\ref{eqn:before-multiply} except with the inequality is flipped:
\begin{align}
\Rightarrow a\frac{-2\natp_1^2}{\natp_2} &\stackrel{?}{\geq}  \sqrt{a} \frac{-2\natp_1^2}{\natp_2} \\
\Rightarrow a &\geq  \sqrt{a} \, ,
\end{align}
where the last step is because we assumed $\natp_2 < 0$ and $a > 1$---note that we do not flip the inequality because $\frac{-2\natp_1^2}{\natp_2}$ is overall a positive number.  Thus, this is an upper bound on the interval $[\bar{\radx}, \infty]$:
\begin{align*}
\A(\natp_1,\natp_2) &= \int_0^{\bar{\radx}} \exp(\natp_1\sqrt{\radx}+\natp_2\radx) \mathrm{d}\radx \\
&\hspace{5em} + \int_{\bar{\radx}}^\infty \exp(\natp_1\sqrt{\radx}+\natp_2\radx) \mathrm{d}\radx \\
&\leq \int_0^{\bar{\radx}} \exp(\natp_1\sqrt{\radx}+\natp_2\radx) \mathrm{d}\radx + \underbrace{\int_{\bar{\radx}}^\infty \exp(\hat{\natp}_2\radx) \mathrm{d}\radx}_{\text{Upper bound}} \\
&\leq \underbrace{\int_0^{\bar{\radx}} \exp(\natp_1\sqrt{\radx}+\natp_2\radx) \mathrm{d}\radx}_{< \infty} + \underbrace{\int_0^\infty \exp(\hat{\natp}_2\radx) \mathrm{d}\radx}_{\text{Exp. log partition}} \\
&< \infty,
\end{align*}
where the last step is based on  the fact that a bounded integral of a finite smooth function is bounded away from $\infty$ and the second term is merely the log partition function of a standard exponential distribution.

\paragraph{Case 3: $\natp_2 = 0$} This gives the log partition function simply as:
\begin{align*}
\A(\natp_1, \natp_2) = \int_0^\infty \exp(\natp_1 \sqrt{\radx}) \mathrm{d}\radx,
\end{align*}
which has the closed form solution:
\begin{align}
&\A(\natp_1, \natp_2) = 2\natp_1^{-2}(\natp_1 \sqrt{\radx} - 1) \exp( \natp_1 \sqrt{\radx}) \Big\rvert_0^\infty \\
&\quad= \lim_{\radx \to \infty} 2\natp_1^{-2}(\natp_1 \sqrt{\radx} - 1) \exp( \natp_1 \sqrt{\radx}) - \left( -2\natp_1^{-2} \right)  \\
&\quad= 2\natp_1^{-2} + \lim_{\radx \to \infty} 2\natp_1^{-2}(\natp_1 \sqrt{\radx} - 1) \exp( \natp_1 \sqrt{\radx})\,. \label{eqn:limit}
\end{align}
The convergence critically depends on the limit in Eqn.~\ref{eqn:limit}.  This limit diverges to $\infty$ if $\natp_1 \geq 0$ but converges to 0 if $\natp_1 < 0$.  Thus, if $\natp_2 = 0$, then $\natp_1 < 0$ for the log partition function to be finite.

\section{Proof of Poisson SQR Normalization (Eqn.~\ref{eqn:poisson-norm})}
First, we take an upper bound by absorbing the $\sqrt{\radx}$ term:
\begin{align}
\A(\natp_1,\natp_2) &\leq \sum_{\radx \in \Zp} \exp\Big( \underbrace{\natp_1 \sqrt{\radx} + \natp_2 \radx}_{O(\radx)} \\
&\hspace{5em}- \underbrace{\sum_\vi \log(\Gamma(\radx\usca_\vi+1))}_{O(\radx \log \radx)} \Big) \\
&\leq \sum_{\radx \in \Zp} \exp\Big( \natp \radx - \sum_\vi \log(\Gamma(\radx\usca_\vi+1)) \Big),
\end{align}
where $\natp = \natp_2 + |\natp_1|$.  We continue the bound as follows:
\begin{align}
&A(\natp_1,\natp_2) \leq \sum_{\radx \in \Zp} \exp\left( \natp \radx - \max_\vi \log(\Gamma(\radx\usca_\vi+1)) \right) \\
& \leq \sum_{\radx \in \Zp} \exp\left( \natp \radx - \log(\Gamma(\radx/\vmax+1)) \right), \label{eqn:vmax}
\end{align}
where Eqn.~\ref{eqn:vmax} comes from the fact that $\arg\max_{\usca_\vi} \log(\Gamma(\radx\usca_\vi + 1)) \geq 1/\vmax$ (simple proof by contradiction).

Now let us use the ratio test for convergent series where $a_\radx = \exp\left( \natp(\uvec) \radx - \log(\Gamma(\frac{\radx}{\vmax}+1)) \right)$:
\begin{align}
&\lim_{\radx \to \infty} \frac{|a_{\radx+1}|}{|a_\radx|} = \exp( \natp(\radx+1) - \log(\Gamma((\radx+1)/\vmax+1)) \notag \\
&\quad\quad\quad - [\natp\radx - \log(\Gamma(\radx/\vmax+1))]) \\
&= \lim_{\radx \to \infty} \exp\left(\natp + \log\left(\frac{\Gamma(\radx/\vmax+1))}{\Gamma((\radx+1)/\vmax+1))} \right) \right) \\
&= \exp(\natp)\lim_{\radx \to \infty} \frac{\Gamma(\radx/\vmax+1))}{\Gamma((\radx/\vmax +1 + 1/\vmax))}\frac{(\radx/p+1)^{1/p}}{(\radx/p+1)^{1/p}} \\
&= \exp(\natp) \lim_{\radx \to \infty} \frac{1}{(\radx/p+1)^{1/p}}  \notag \\
&\hspace{6em} \times \lim_{\radx \to \infty} \frac{\Gamma(\radx/\vmax+1))(\radx/p+1)^{1/p}}{\Gamma((\radx/\vmax +1 + 1/\vmax))} \label{eqn:limit-product} \\
&= \exp(\natp) \lim_{\radx \to \infty} \frac{1}{(\radx/p+1)^{1/p}} (1) \label{eqn:gamma-limit}\\
&= \exp(\natp) (0)(1) = 0 < 1,
\end{align}
where Eqn.~\ref{eqn:limit-product} is by the product of limits rule and Eqn.~\ref{eqn:gamma-limit} is by the well-known asymptotic properties of gamma functions.  Therefore, by the ratio test, the radial conditional log partition function is bounded for any $\natp_1 < \infty$ and $\natp_2 < \infty$.

\end{document}